\documentclass[conference,9pt]{IEEEtran}
\IEEEoverridecommandlockouts

\usepackage{cite}
\usepackage{amsmath,amssymb,amsfonts}
\usepackage{algorithmic}
\usepackage{graphicx}
\usepackage{textcomp}
\usepackage{xcolor}

\usepackage{booktabs} 
\usepackage{multirow}
\usepackage{makecell}
\usepackage{wrapfig}

\def\BibTeX{{\rm B\kern-.05em{\sc i\kern-.025em b}\kern-.08em
    T\kern-.1667em\lower.7ex\hbox{E}\kern-.125emX}}
\begin{document}


\title{Generating Editable Head Avatars with 3D Gaussian GANs\\
\thanks{
This work is partly supported by the National Key R\&D Program of China (2022ZD0161902), the National Natural Science Foundation of China (No. 62202031), Beijing Municipal Natural Science Foundation (No. 4222049), and the Fundamental Research Funds for the Central Universities.

Corresponding author: Yunhong Wang.

© 20XX IEEE.  Personal use of this material is permitted.  Permission from IEEE must be obtained for all other uses, in any current or future media, including reprinting/republishing this material for advertising or promotional purposes, creating new collective works, for resale or redistribution to servers or lists, or reuse of any copyrighted component of this work in other works.
}
}


\author{
    \IEEEauthorblockN{\em Guohao Li$^1$, Hongyu Yang$^{2,3}$, Yifang Men, Di Huang$^1$, Weixin Li$^{1,3}$, Ruijie Yang$^1$ and Yunhong Wang$^1$}

    \IEEEauthorblockA{\\
    $^1$ School of Computer Science and Engineering, Beihang University, Beijing, China\\
    $^2$ School of Artificial Intelligence, Beihang University, Beijing, China\\
    $^3$ Shanghai Artificial Intelligence Laboratory, Shanghai, China
    }
}

\maketitle

\begin{abstract}
Generating animatable and editable 3D head avatars is essential for various applications in computer vision and graphics.
Traditional 3D-aware generative adversarial networks (GANs), often using implicit fields like Neural Radiance Fields (NeRF), achieve photorealistic and view-consistent 3D head synthesis. However, these methods face limitations in deformation flexibility and editability, hindering the creation of lifelike and easily modifiable 3D heads. 
We propose a novel approach that enhances the editability and animation control of 3D head avatars by incorporating 3D Gaussian Splatting (3DGS) as an explicit 3D representation.
This method enables easier illumination control and improved editability.
Central to our approach is the Editable Gaussian Head (EG-Head) model, which combines a 3D Morphable Model (3DMM) with texture maps, allowing precise expression control and flexible texture editing for accurate animation while preserving identity.
To capture complex non-facial geometries like hair, we use an auxiliary set of 3DGS and tri-plane features. 
Extensive experiments demonstrate that our approach delivers high-quality 3D-aware synthesis with state-of-the-art controllability.
Our code and models are available at https://github.com/liguohao96/EGG3D.
\end{abstract}


\newcommand{\lowb}{$\downarrow$}  
\newcommand{\highb}{$\uparrow$}   
\newcommand{\ie}{\textit{i.e.}}  
\newcommand{\eg}{\textit{e.g.}}  

\definecolor{Gray}{gray}{0.9}
\definecolor{LightGray}{gray}{0.6}
\newcommand*{\cg}{\color{LightGray}}

\section{Introduction}

Generating animatable and editable 3D heads is increasingly vital for applications in virtual reality, video synthesis, and gaming. 
The field has seen rapid growth, with numerous promising methods emerging \cite{eg3d,rodin,ssif,rgca,anifacegan,next3d,aniportraitgan,icassp_adaptive,icassp_e2e_gen,icassp_meta_talk,icassp_3dmm_diff}. 
Among these techniques, animatable 3D-aware Generative Adversarial Networks (GANs) \cite{gnarf,next3d} have garnered significant attention for their ability to rapidly generate photo-realistic videos and 3D geometries.
Typically, these methods utilize neural implicit fields as the 3D representation and incorporate 3D Morphable Models (3DMM) \cite{flame} for animatable conditions (e.g., expressions).
However, the lack of deformation flexibility and editability in neural implicit fields hinders accurate animation control and the ability to edit the generated results \cite{gnarf,neumesh,mega}.


3D Gaussian Splatting (3DGS) \cite{3dgs} is a recently introduced 3D representation based on point clouds, showing significant potential for real-time rendering of photorealistic images. Its ability to capture fine geometries \cite{2dgs}, combined with inherent flexibility, makes it well-suited for tasks involving deformation and editing \cite{physgaussian}. These characteristics position 3DGS as a promising approach for generating 3D head avatars.
However, integrating 3DGS into a 3D-aware generative model presents significant challenges:
1) Ambiguities between Gaussian point scales and distances to the camera.
For instance, a point near the camera with a small scale can be visually equivalent to a point farther from the camera with a larger scale after rasterization, which can lead to issues like floaters.
This ambiguity becomes even more problematic within a 3D-aware generation framework, where the absence of supervision from other views exacerbates the problem.
2) Unstructured nature of point clouds poses challenges in achieving the generalizability and structured representation required by GAN.

In this study, we introduce a novel method, \textbf{E}ditable \textbf{G}aussian \textbf{G}eometry \textbf{3D} GAN (EG$^2$3D), for generating animatable and editable 3D heads by leveraging the efficiency and flexibility of 3DGS. 
To address the ambiguities and the unstructured nature mentioned earlier, we decouple the learning of position and attributes for Gaussian points using 3DMM incorporating with well-structured representations.
This approach enables the generation of animatable and editable 3D heads within the framework of 3D-aware GANs.
We propose that the intricate variations in geometry across individuals can be accurately captured by a dense, shared point cloud in space, which is distinguished by varying ``visibility'' attributes. The opacity attribute of 3DGS serves as an effective representation for the visibility of points, thereby preventing 3D Gaussians from drifting freely and causing artifacts like floaters.


Given the structured nature of the human facial region \cite{flame} and the complex geometry of non-facial elements such as hair and eyeglasses, we propose a partitioning strategy within our shared point cloud. 
Specifically, we divide the point cloud into two distinct groups. 
For the \textit{face region}, we introduce the \textbf{E}ditable \textbf{G}aussian Head model (EG-Head), which integrates the animatable FLAME model \cite{flame} with editable texture maps in a structured UV space. This model employs a coarse-to-fine generation process, producing a fixed number of Gaussian points with appearances derived from sampling mesh and texture maps, ensuring accurate expression animation and identity consistency.
Meanwhile, to address the intricate geometry of \textit{non-facial regions} such as hair, we employ a Gaussian point cloud with restricted spatial variation, complemented by a structured tri-plane representation.

In the proposed framework, these two groups of point clouds are generated collaboratively and rendered simultaneously using 3DGS \cite{3dgs}. By learning different visibility attribute alongside other appearance attributes, our method ensures significant texture and geometry variations (e.g., different hairstyles), enhancing the diversity of generated outputs. Additionally, illumination control is achieved through Spherical Harmonic (SH) lighting and a lightweight CNN that predicts shadowing effects based on rendered buffers.
To stabilize the training process, we utilize region-based masks in the discriminator, ensuring that the Gaussian point cloud is accurately generated.

Our contributions can be summarized as:
\label{intro_contribution}
\begin{itemize}
\item We propose EG$^{2}$3D, a novel method for generating 3D head avatars, enabling highly controllable and animatable 3D head generation within a 3D-aware GAN framework.
\item We introduce the EG-Head model, an advanced head representation that integrates the FLAME model with texture maps, enabling precise animation and flexible editing of facial expressions and localized attributes. 
\item EG$^{2}$3D demonstrates significant improvements in animation accuracy and identity preservation, representing a substantial advancement in 3D head avatar generation.
\end{itemize}


 \begin{figure*}
    \centering
    \resizebox{0.9\linewidth}{!}{
    \includegraphics{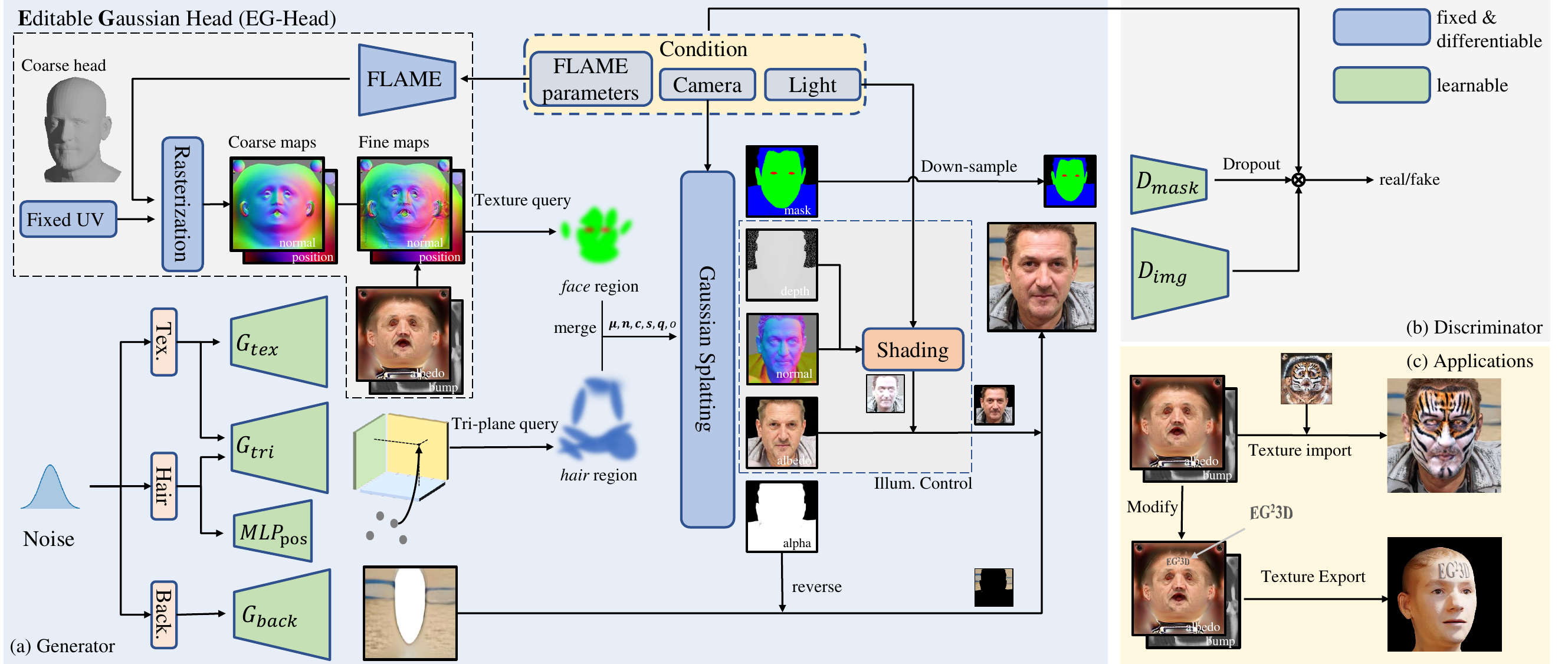}
    }
    \caption{Overview of the proposed EG$^2$3D method. (a) \textbf{The generator:} \(G_{tex}\) generates texture maps for the \textbf{E}ditable \textbf{G}aussian Head (EG-Head) representing the face region. \(G_{tri}\) and \(\text{MLP}_{pos}\) generate a Gaussian point cloud with tri-plane for the hair region, and a background generator \(G_{back}\) is separately built. These Gaussian point clouds are merged and rendered directly at high resolution through Gaussian splatting with injected illumination information.
  (b) \textbf{The discriminators:} Separate discriminators \(D_{mask}\) and \(D_{img}\) help to learn the position and appearance of each Gaussian point clouds. (c) \textbf{Applications with EG$^2$3D:} Rendering with modified texture and exporting to 3D assets.
    }
    \label{fig:framework}
    \vspace{-0.6cm}
\end{figure*}

\section{Method}
\label{method}

This paper introduces \textbf{EG$^2$3D}, a novel 3D-aware GAN framework for generating animatable and editable 3D head avatars. Our approach leverages 3DGS and is conditioned on identity, expression, lighting, and camera parameters to achieve high fidelity and controllability. 
To address the inherent ambiguity in 3DGS, we propose generating a shared point cloud that not only supports animatability and editability but also encapsulates diverse attributes. 
To stabilize the distribution of these Gaussian points, we employ a hybrid approach that generates a 3D-aware foreground using 3DGS and a 2D image generator for the background.

We extend the standard 3DGS framework \cite{3dgs} by incorporating an auxiliary normal direction, defined as the \(z\)-axis of the rotation matrix, to support illumination.
Each Gaussian point is characterized by its center position \(\mathbf{\mu} \in \mathbb{R}^3\), normal direction \(\mathbf{n} \in \mathbb{R}^3\), color \(\mathbf{c} \in \mathbb{R}^3\), scale \(\mathbf{s} \in \mathbb{R}^3\), quaternion \(\mathbf{q} \in \mathbb{R}^4\), and opacity \(\mathbf{o} \in \mathbb{R}\).
We further segment the shared point cloud into two distinct groups, one representing the relatively structured \textit{face} region and the other capturing the intricate geometry of the \textit{hair} region, ensuring that each region is accurately modeled according to its unique geometric characteristics, enabling precise representation of the complex foreground head from 2D image data.

For the \textit{face} region, we introduce the \textbf{Editable Gaussian Head (EG-Head)}, which integrates the 3D Morphable Model (3DMM) and structured texture maps. This facilitates precise animation and editing of facial geometry, while preserving the identity of the subject. For the non-facial \textit{hair} region, we employ a Gaussian point cloud with limited spatial variation, complemented by attributes derived from a structured tri-plane representation \cite{eg3d}. 
These two groups of Gaussian point clouds are then merged and rendered at high resolution using Gaussian splatting \cite{3dgs}. 
To maintain efficient rendering, 
we implement illumination control using screen-space buffers combined with spherical harmonics (SH) lighting and a lightweight CNN to model shadowing effects.
In the subsequent sections, we provide a detailed exposition of our framework.

\newcommand{\resimg}{{t_{img}}}  
\newcommand{\restex}{{t_{tex}}}  
\newcommand{\restri}{{t_{tri}}}  
\newcommand{\resseg}{{t_{mask}}} 

\newcommand{\alphead}{{\alpha_{face}}} 
\newcommand{\alphair}{{\alpha_{hair}}} 

\newcommand{\headp}{\mathbf{\mu}^{face}}
\newcommand{\headn}{\mathbf{n}^{face}}
\newcommand{\headc}{\mathbf{c}^{face}}
\newcommand{\heads}{\mathbf{s}^{face}}
\newcommand{\headq}{\mathbf{q}^{face}}
\newcommand{\heado}{\mathbf{o}^{face}}

\newcommand{\hairp}{\mathbf{\mu}^{hair}}
\newcommand{\hairn}{\mathbf{n}^{hair}}
\newcommand{\hairc}{\mathbf{c}^{hair}}
\newcommand{\hairs}{\mathbf{s}^{hair}}
\newcommand{\hairq}{\mathbf{q}^{hair}}
\newcommand{\hairo}{\mathbf{o}^{hair}}

\newcommand{\joinp}{\mathbf{\mu}^{join}}
\newcommand{\joinn}{\mathbf{n}^{join}}
\newcommand{\joinc}{\mathbf{c}^{join}}
\newcommand{\joins}{\mathbf{s}^{join}}
\newcommand{\joinq}{\mathbf{q}^{join}}
\newcommand{\joino}{\mathbf{o}^{join}}

\subsection{Editable Gaussian Head Generation}

We propose the Editable Gaussian Head, which facilitates animation and editing by utilizing the FLAME model \cite{flame} along with a generated albedo map \( A_{tex} \in \mathbb{R}^{\restex \times \restex \times 3} \) and bump map \( B_{tex} \in \mathbb{R}^{\restex \times \restex \times 1} \) with resolution \( \restex^2 \). EG-Head disentangles the position and attributes of the Gaussian point cloud, providing intuitive editing capabilities, as shown in Fig.~\ref{fig:framework}. This is achieved by incorporating the albedo map \( A_{tex} \) as a static appearance texture map for the color attribute \( \mathbf{c} \). The bump map \( B_{tex} \) is utilized as a geometry texture map to represent fine-grained geometry by offsetting the FLAME mesh surface along its normal.

As shown in Fig.\ref{fig:framework}(a), we initially generate a coarse head mesh using the provided conditions with a frozen FLAME model. The vertex positions of the coarse head mesh, along with its surface normals, are projected into UV space with resolution \(\restex\) through differentiable rasterization \cite{nvdiffrast}, producing a coarse position map \( P^C_{tex} \in \mathbb{R}^{\restex \times \restex \times 3} \) and a coarse normal map \( N^C_{tex} \in \mathbb{R}^{\restex \times \restex \times 3} \).

The detailed position map \( P^F_{tex} \in \mathbb{R}^{\restex \times \restex \times 3} \) is generated using the bump map \( B_{tex} \) as follows: \( P^F_{tex} = P^C_{tex} + B_{tex} \cdot N^C_{tex} \). By computing the discrete differentials of \( P^F_{tex} \) along the \( u \)-axis and \( v \)-axis in UV space, denoted as \( \Delta_u P^F_{tex} \) and \( \Delta_v P^F_{tex} \), the detailed normal map \( N^F_{tex} \in \mathbb{R}^{\restex \times \restex \times 3} \) is obtained by: \( N^F_{tex} = \Delta_u P^F_{tex} \times \Delta_v P^F_{tex} \).
Next, the color $\headc_k$, center position $\headp_k$ and normal direction $\headn_k$ for the $k$-th Gaussian point in the \textit{face} region are sampled from $A_{tex}$, ${P^F_{tex}}$ and ${N^F_{tex}}$ respectively through texture query.
Since the face region is relatively flat, we assume that the Gaussian points are always visible and are as thin as a \textit{2D disk}.
We use a relatively small value \(\varepsilon\) as the fixed scale of the \( z \)-axis and two learnable scalar for x and y axis. 
Quaternion $\headq_k$ is computed from normal $\headn_k$.
In practice, we generate a fixed number of \( \#face = \alpha_{\text{head}} \times \restex \times \alpha_{\text{head}} \times \restex \) Gaussian points, where \( \alpha_{\text{head}} \) is a hyper-parameter for the sampling ratio. The position \( \headp_k \) of each Gaussian point is sampled from the dynamic position map \( P^F_{\text{tex}} \), which accurately animates along with the coarse FLAME mesh. The color attribute \( \headc_k \) of each Gaussian point is sampled from a static albedo map with fixed coordinates \((u_k,v_k)\). This approach facilitates a disentangled representation for position and appearance, ensuring that only the position changes during animation while the appearance remains fixed, thus helping to preserve identity during animation.
Since color and detailed geometry are controlled by 2D texture maps  \( A_{tex} \) and  \( B_{tex} \), EG-Head can faithfully render any edits within these texture maps, as shown in Fig.~\ref{fig:framework}(c).
These texture maps are well-structured and can be easily learned with 2D CNNs, thanks to the consistent 2D texture space.


\subsection{Tri-Plane Gaussian Representation for Hair Region}
To ensure a faithful representation of the sophisticated \textit{hair} region, we propose a generalizable Gaussian point cloud with tri-plane feature maps with resolution \(\restri\)$^2$. We generate a fixed number of \( \#hair = \alpha_{\text{hair}} \times \restri \times \alpha_{\text{hair}} \times \restri \) Gaussian points for the hair region, where the hyper-parameter \( \alpha_{\text{hair}} \) controls the number of points. We assume that the position \( \mathbf{\mu}^{\text{hair}} \) of the Gaussian points is shared in order to address the ambiguity of 3DGS and stabilize training. Thus, the Gaussian point cloud will receive supervision from other views and enforce learning of varying tri-plane.

As shown in Fig.~\ref{fig:framework}(a), the shared Gaussian point cloud \( \hairp \in \mathbb{R}^{\#hair \times 3} \) is generated from \( {MLP}_{\text{pos}} \) with additional conditions on identity and expression. 
We modify the output layer of \({MLP}_{\text{pos}}\) as \( \text{tanh}(W^Tx) + b \), such that the generated point cloud has little spatial variation around the \textit{bias} \( b \in \mathbb{R}^{\#hair \times 3} \).
Tri-plane features for the \( k \)-th Gaussian point are sampled from the generated tri-plane feature maps and decoded into color $\hairc_k$, normal $\hairn_k$, scale $\hairs_k$, quaternion $\hairq_k$ and opacity $\hairo_k$, with a decoding MLP. 


\subsection{Integrating 3DGS with GAN Training}
\textbf{3DGS with Light Condition}.
As shown in Fig.~\ref{fig:framework}(a), the Gaussian point clouds for the \textit{face} region and \textit{hair} region are merged into a single Gaussian point cloud represented by \(\joinp\), \(\joinn\), \(\joinc\), \(\joins\), \(\joinq\), and \(\joino\). 
We perform 3D Gaussian splatting with a modified ``color'' channel by concatenating group ID, depth, \(\joinn\), and \(\joinc\).
Here, the group ID consists of three channels (red, green and blue for eye, face and hair, respectively), and depth is calculated as the distance between the camera center and the position \(\joinp_k\).
Subsequently, we separate the rendered ``color'' into a mask, depth buffer, normal buffer, albedo buffer, and alpha buffer, with a resolution of \(\resimg\)$^2$.
To faithfully and accurately control lighting, we propose injecting illumination conditions using these intermediate rendering buffers with SH lighting and a lightweight CNN which predicts a \textit{occlusion} factor for each pixel through comparing neighboring depths based on the normal direction to determine whether light is occluded.

\textbf{Discriminator}. Our discriminator is based on the architecture proposed in EG3D \cite{eg3d}.
Since our approach directly renders at high-resolution $\resimg^2$, the input to the discriminator is the RGB image along with the three channel masks.
Real masks are pre-computed using off-the-shelf face parsing network \cite{farl}.
We utilize two separate discriminator backbones: $D_{img}$ with resolution $\resimg^2$ for the RGB images, and $D_{mask}$ with resolution $\resseg^2$ for masks.
This setup helps to stabilize the training and aids in learning the position and appearance of these two group of Gaussian point clouds for foreground.
This design is referred to as the \textit{Separate Discriminator}.

\textbf{Regularization}. 
Similar to EG3D, we utilize the R1 penalty to regularize the discriminator, with separate hyper-parameters, \(\gamma_{\text{mask}}\) and \(\gamma_{\text{img}}\), for masks and RGB images. 
A smoothness constraint is applied to the generated texture maps and background. Additionally, we enforce the generated albedo map to be as symmetrical as possible and penalize the bump map with the L1 norm.


\section{Experiments}
\label{experiment}


\subsection{Dataset and Experimental Setup}


\begin{table}[]
    \caption{Comparison with SoTA 3D-aware GANs. }
    \centering
    \resizebox{0.49\textwidth}{!}{
    \begin{tabular}{ccccccc}
        \toprule
                                 Method                  & FID\lowb     & AED\lowb & APD\lowb & AID\lowb & ID\highb & FPS\highb \\
        \midrule
                                Next3D \cite{next3d}     & $3.9^\dag$   & 0.3087 & 0.0483 & \cg{0.3005} & 0.7200 & 13 \\
                                EG3D \cite{eg3d}         & $4.7^\dag$   & \cg{0.3647} & 0.0509 & \cg{0.3008} & 0.6987 & 35* \\
                                Trevithick et al. \cite{whatyousee} & 4.97$^\dag$  & - & -          & - & - & 4.5*$^\dag$\\
                                Ours                    & 7.51         & \bf{0.2347} & \bf{0.0410} & \bf{0.1556} & \bf{0.7731} & \bf{65}* \\
        \bottomrule
        \multicolumn{7}{l}{
        $\ ^\dag$ indicates results from their original paper. $*$ indicates using cache.}
    \end{tabular}
    \label{tab:compare}
    }
\end{table}

\textbf{Dataset}. We train and test the proposed EG$^2$3D method on the FFHQ dataset \cite{stylegan} with resolution $512^2$. 
To enhance the dataset, we apply horizontal flips and use the EMOCA v2 \cite{emocav2} 3D face reconstruction model to predict camera pose, FLAME parameters (including identity, expression, jaw pose, and eye poses), as well as SH lighting coefficients. Since EMOCA v2 employs an orthogonal camera projection model, we convert it to a pinhole camera with constant intrinsics for better realism. To achieve more accurate camera poses and greater control over eyeball movement, we use these parameters as an initial estimate and further refine the FLAME model to fit the images. This fitting process integrates landmarks predicted by \cite{mediapipe} and face parsing results from \cite{farl}. The face parsing outputs are then converted into masks, which are fed into $D_{mask}$ to stabilize the positioning of the Gaussian point cloud.



\subsection{Comparison}


\begin{figure}
    \centering
    \resizebox{0.99\linewidth}{!}{
    \includegraphics[page=1,trim=40 40 0 0]{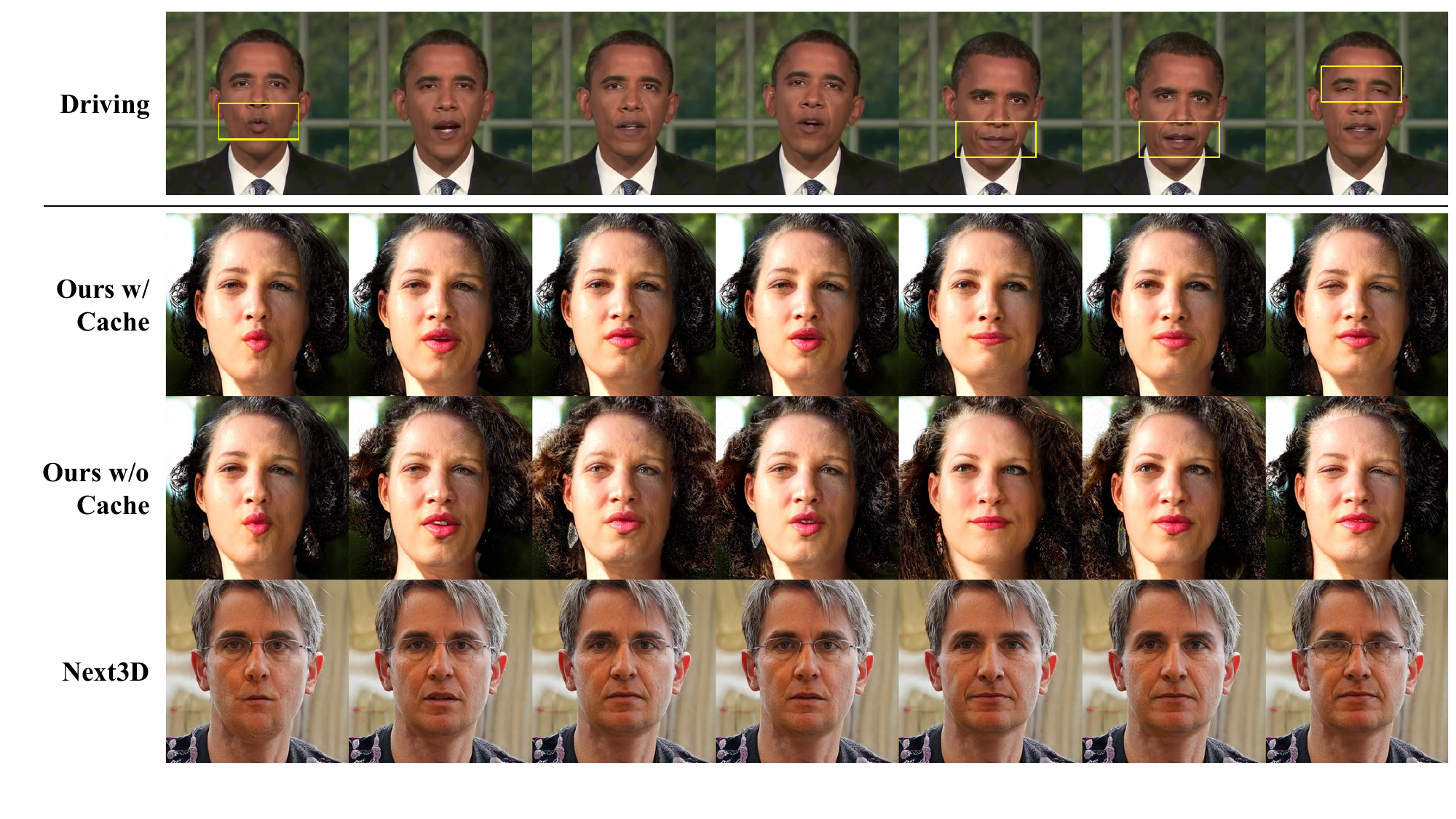}
    }
    \caption{Comparison with state-of-the-art animatable 3D GANs on video driving. 
    }
    \label{fig:driving}
    \vspace{-0.4cm}
\end{figure}

\textbf{Quantitative comparison}.
We measure image quality with Fréchet Inception Distance (FID) \cite{fid}.
We evaluate the controllability using Average Expression Distance (AED) and Average Pose Distance (APD) following \cite{gnarf, next3d}, additionally introduce Average Illumination Distance (AID). 
Following EG3D \cite{eg3d}, We evaluate identity consistency (ID) across generated multi-view images.
As shown in Tab.~\ref{tab:compare}, our method achieves superior scores in animation control (AED and APD), illumination control (AID) and identity consistency (ID)  compared to all other methods at a real-time rendering speed.
These results underscore the substantial potential of our approach for various downstream tasks.
In terms of the quality of generated images, our method attains an FID score of 7.51. 
The reduction in image quality observed in our approach can be attributed to the inherent trade-off between controllability and visual fidelity, a challenge also noted in prior works such as \cite{hybrid_generator, nerffaceediting}.
The introduction of albedo-lighting disentanglement for illumination control inherently poses a challenge to maintaining high image quality.
The absence of ground-truth HDR environment maps and precise geometry necessitates relying on prior models to approximate lighting and shadowing.
However, this trade-off is necessary to achieve the desired level of control over the generated content, even though it may slightly diminish the realism of the rendered images.
Our approach prioritizes these controllable features, accepting minor compromises in image quality.

\begin{figure}
    \centering
    \resizebox{0.99\linewidth}{!}{
    \includegraphics{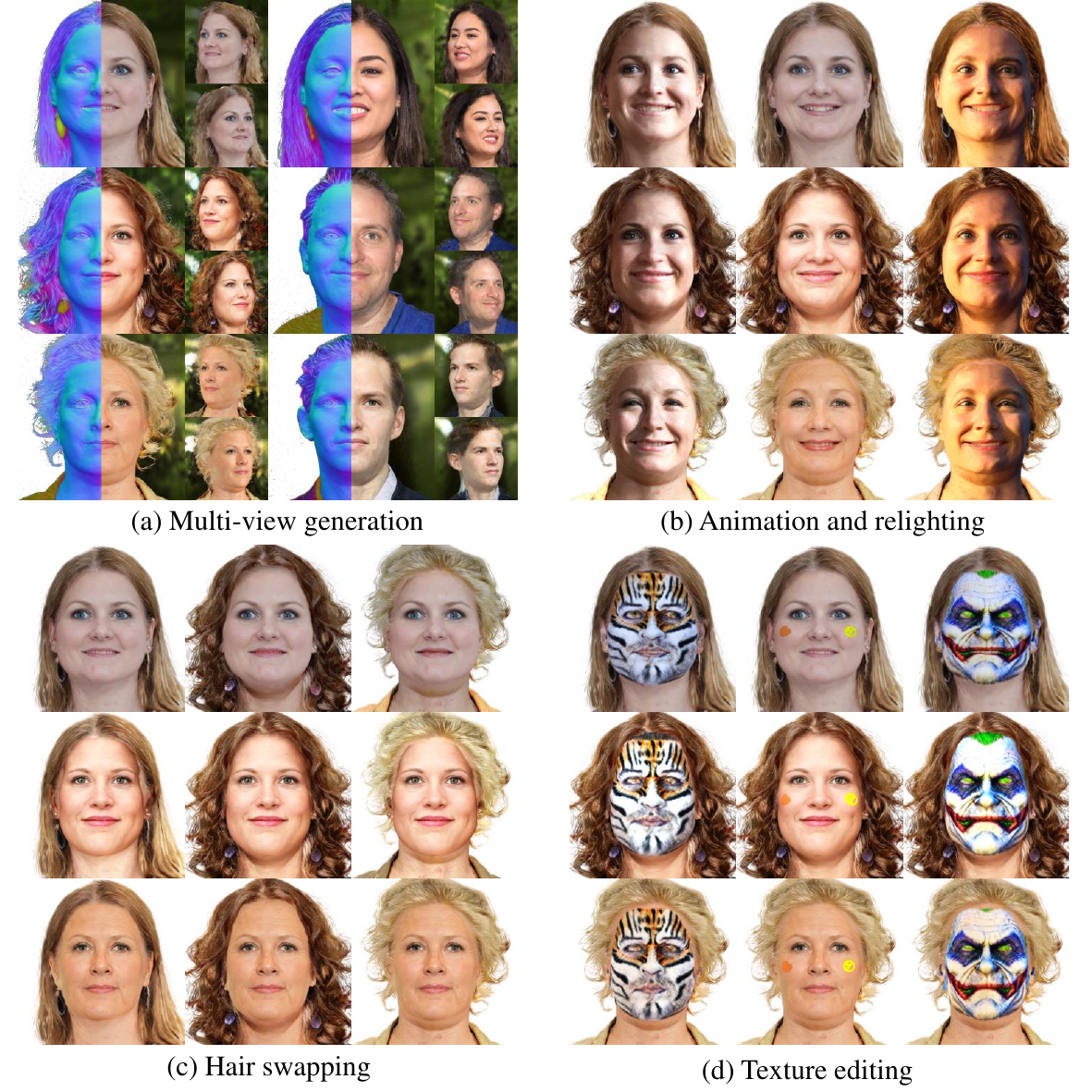}
    }
    \caption{Generated animatable and editable 3D avatars. Editing textures are courtesy of \cite{wsdf}.}
    \label{fig:application}
    \vspace{-0.2cm}
\end{figure}

\textbf{Qualitative comparison}.
We compare the animation capabilities of our method with Next3D by driving both methods using estimated FLAME parameters for expression and jaw pose derived from a video sequence via EMOCA v2 \cite{emocav2}. As shown in Fig.~\ref{fig:driving}, our method (w/o Cache) achieves precise expression driving results, particularly in the mouth and eye regions.
Additionally, our method benefits from disentangled position and attributes with the proposed EG-Head, enabling us to cache the attributes of 3DGS and achieve expression driving through position deformation along with FLAME parameters. We further generate driving results using our cache mechanism (w/ Cache), where attributes generated from the first image are cached for subsequent images. As shown in Fig.~\ref{fig:driving}, applying cached Gaussian attributes ensures superior identity preservation without compromising the accuracy of expression driving.

\subsection{Ablation Study}
We performed an ablation study on the proposed EG-Head model, which is designed to enable precise expression control and flexible texture editing while preserving identity.
We begin with a baseline model where no texture or geometry maps are used (denoted as ``w/o maps'' in Table~\ref{tab:ablation}). Next, we augment the model by adding the appearance map, without utilizing bump maps (denoted as ``+ appearance tex.'' in Table~\ref{tab:ablation}). This modification significantly improves identity consistency (ID) and illumination controllability (AID), with only a minor impact on generation quality (FID).
We then extend the model by incorporating the geometry map, forming the complete EG-Head (denoted as ``+ geometry map'' in Table~\ref{tab:ablation}). The addition of the geometry map results in a marked improvement in both generation quality (FID) and identity consistency (ID), further demonstrating the effectiveness of the proposed approach.

\begin{table}[!h]
    \centering
    \vspace{-0.5em}
    \caption{Ablation study results. Models are trained at a resolution of \(128^2\) with 5 million images.}
    \label{tab:ablation}
    \resizebox{0.95\linewidth}{!}{
    \begin{tabular}{lccccc}
        \toprule
        Method                 & FID\lowb   & AED\lowb    & APD\lowb    & AID\lowb    & ID\highb \\
        \midrule  
        w/o maps                       & 16.79      & 0.2982      & 0.0400      & 0.2300      & 0.7005 \\
        + appearance map               & 18.46      & \bf{0.2685} & 0.0400      & 0.1689      & 0.7582 \\
        + geometry map (Ours $128^2$)  & \bf{15.51} & 0.2837      & \bf{0.0397} & \bf{0.1688} & \bf{0.7731} \\
        \bottomrule
    \end{tabular}
    }
    \vspace{-1.5em}
\end{table}

\subsection{Applications}
We demonstrate the generated animatable and editable 3D avatar in Fig. \ref{fig:application}.
First, we generate 3D avatars, and render with different yaw angles, synthesising identity-consistency multi-view images, as shown in Fig. \ref{fig:application}(a).
Next, the animation capabilities of the generated 3D avatar are illustrated by altering the expression, eye pose, and illumination, as shown in Fig. \ref{fig:application}(b).
We further demonstrate our approach’s ability to swap hair styles by exchanging the \textit{hair} Gaussian point clouds of different avatars, as shown in Fig. \ref{fig:application}(c).
Finally, we demonstrate the editability of our approach by editing on the texture maps, as shown in Fig. \ref{fig:application}(d),
This further demonstrates the huge potential of our approach in downstream tasks.
Please refer to our code base for synthesised video.

\section{Conclusion}
This paper proposes EG$^2$3D, a 3D-aware GAN framework that incorporates 3DGS to enable the training of editable 3D head generation from a corpus of 2D images.
The proposed EG-Head 
enables precise animation with consistent identity and facilitating inversion-free editability.
Our approach directly renders at high resolution on-the-fly without super resolution, leveraging an illumination injection module. 
Experimental results demonstrate that EG$^2$3D effectively learns an editable 3D head with significantly improved animation accuracy and identity consistency. Our work paves the way for more advanced 3D head generation using 3DGS.

\section{Discussions}
\begin{wrapfigure}{r}{0.3\linewidth}
    \vspace{-1em}
    \includegraphics[page=1,width=1.0\linewidth,trim=0 150 0 0]{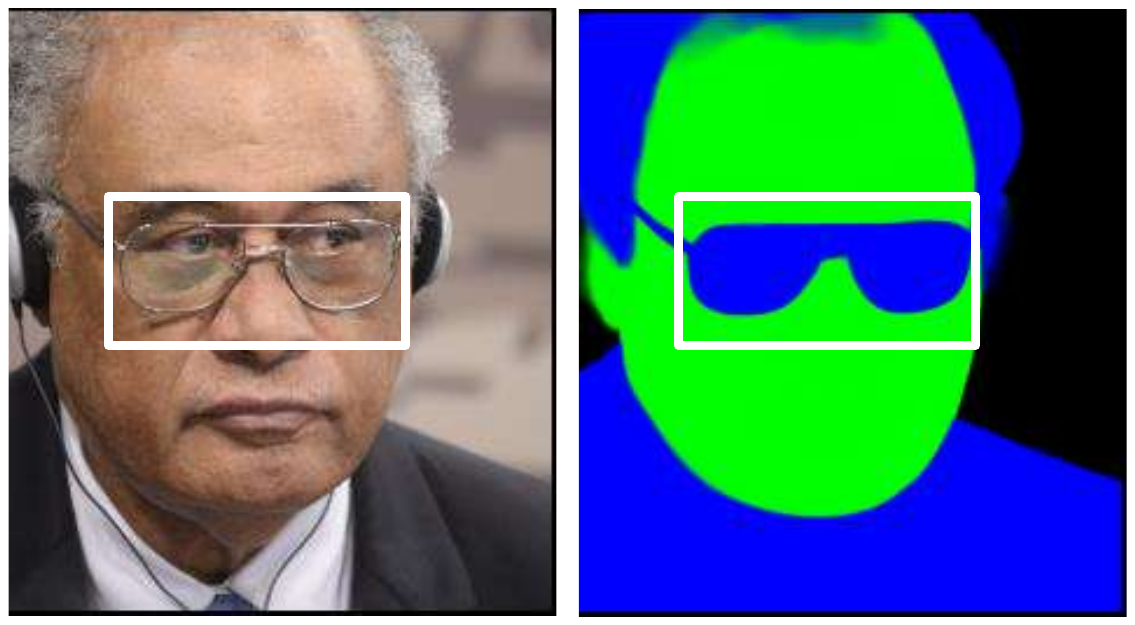}
    \caption{}
    \vspace{-1em}
    \label{fig:failure}
\end{wrapfigure}
\textbf{Failure Case}.
Our method failed to generate transparent eyeglasses due to the bias of face parsing network, as illustrated in Fig.~\ref{fig:failure}.
The face parsing network predicts an opaque mask for eyeglass, which should be opacity.

\textbf{Future works}
In this work, we have demonstrated the potential of using 3D Gaussian Splatting (3DGS) for generating animatable and editable 3D avatars with improved identity consistency, expression control, and texture map editability. 
There remain several avenues for future improvements and exploration.

One potential direction is to explore the integration of larger and more diverse datasets to mitigate the current limitations imposed by biases in the data (\eg limited view angles, biased eyeglass opacity).
Additionally, the framework could also benefit from continuous improvements in 3DMM models and more advanced generation architectures that can better learn high-quality geometry and appearance from large corpus of 2D data.

Another area for enhancement is the modeling of body regions, coupled with the Physically Based Rendering (PBR) techniques.
While our method currently focuses on the head region, extending the framework for full-body generation presents an exciting avenue for creating comprehensive avatars.
Moreover, applying PBR could further improve the realism of the renderings, allowing for more photorealistic and lifelike animations.
This would ultimately expand the framework’s potential for creating fully integrated, editable avatars, with diverse applications in virtual environments and gaming.


\bibliographystyle{IEEEtran}
\bibliography{
    icassp,
    reference/bib_main,
    reference/bib_3dgen,
    reference/bib_graphic,
    reference/bib_gan,
    reference/bib_data
}

\end{document}